\pdfoutput=1

\documentclass[11pt]{article}

\usepackage[preprint]{acl}

\usepackage{times}
\usepackage{latexsym}

\usepackage[T1]{fontenc}

\usepackage[utf8]{inputenc}

\usepackage{microtype}

\usepackage{inconsolata}

\usepackage{graphicx}
\usepackage{tabularx}
\usepackage{multirow}
\usepackage{booktabs}
\usepackage{tablefootnote}
\usepackage{verbatim}
\usepackage{dingbat}
\usepackage{soul}
\usepackage{kotex}
\usepackage{amsmath}
\usepackage{subcaption}
\usepackage[normalem]{ulem}
\useunder{\uline}{\ul}{}

\newcommand{\sy}[1]{{\color{blue}#1}}
\usepackage{graphicx}
\usepackage{authblk}

\newcommand\correspondingauthor{\textsuperscript{\ddag}} 
\newcommand\equalcontributor{\textsuperscript{*}} 
\newcommand\current{\textsuperscript{\textdagger}}              

\begin{document}

\title{SALAD: Improving Robustness and Generalization through Contrastive Learning with Structure-Aware and LLM-Driven Augmented Data}

\author{
 \textbf{Suyoung Bae\textsuperscript{1}\equalcontributor},
 \textbf{Hyojun Kim\textsuperscript{2}\equalcontributor\current},
 \textbf{YunSeok Choi\textsuperscript{1}\correspondingauthor},
 \textbf{Jee-Hyong Lee\textsuperscript{1}\correspondingauthor} \\
 \textsuperscript{1} Sungkyunkwan University, South Korea \\
 \textsuperscript{2} SK Telecom, South Korea \\
 \textsuperscript{1} \{sybae01, ys.choi, john\}@skku.edu, 
 \textsuperscript{2} hjkim@sk.com
}
\maketitle


\renewcommand{\thefootnote}{\fnsymbol{footnote}} 
\footnotetext[1]{These authors contributed equally to this work.}          
\footnotetext[2]{Work was conducted during his graduate studies at Sungkyunkwan University.} 
\footnotetext[3]{Co-corresponding authors.}       
\renewcommand{\thefootnote}{\arabic{footnote}} 

\begin{abstract}
In various natural language processing (NLP) tasks, fine-tuning Pre-trained Language Models (PLMs) often leads to the issue of spurious correlations, which negatively impacts performance, particularly when dealing with out-of-distribution data.
To address this problem, we propose \textbf{SALAD} (\textbf{S}tructure \textbf{A}ware and \textbf{L}LM-driven \textbf{A}ugmented \textbf{D}ata), a novel approach designed to enhance model robustness and generalization by generating structure-aware and counterfactually augmented data for contrastive learning.
Our method leverages a tagging-based approach to generate structure-aware positive samples and utilizes large language models (LLMs) to generate counterfactual negative samples with diverse sentence patterns. By applying contrastive learning, \textit{SALAD} enables the model to focus on learning the structural relationships between key sentence components while minimizing reliance on spurious correlations.
We validate our approach through experiments on three tasks: Sentiment Classification, Sexism Detection, and Natural Language Inference. The results demonstrate that \textit{SALAD} not only improves model robustness and performance across different environments but also enhances generalization to out-of-distribution datasets and cross-domain scenarios.
\end{abstract}

\section{Introduction}

In many natural language processing (NLP) tasks, machine learning models often suffer from the issue of spurious correlations (a.k.a shortcuts) between input text tokens and output labels~\citep{tu-etal-2020-empirical}.
These shortcuts allow models to rely on irrelevant patterns in the data, leading to biased predictions. 
For example, a model trained mainly on positive reviews of \textit{Spielberg} movies may incorrectly associate the word \textit{Spielberg} with favorable sentiment, regardless of the actual content of the review~\citep{wang-culotta-2020-identifying, wang-etal-2022-identifying}.
This reliance on superficial patterns results in poor performance, especially when handling out-of-distribution data. 

Many research has addressed this problem by exploring methods for generating counterfactually augmented data (CAD)~\citep{Kaushik2020Learning, samory2021call}, intended to disrupt these false correlations. 
The CAD method involves altering input data to flip its label, providing models with examples that help overcome shortcut-based learning and improve generalization across diverse datasets. However, previous approaches have focused mainly on manual generation methods and statistical techniques to automate CAD creation. Although manual methods can produce high-quality counterfactual data, they are both costly and time-consuming. 

To reduce these costs and improve scalability, automated techniques have been developed, such as using sentiment dictionaries~\citep{yang-etal-2021-exploring}, statistical matching, or predefined antonyms~\citep{Wang_Culotta_2021}. However, these methods are dependent on fixed rules that restrict the quality of the generated data. For example, sentiment dictionaries and statistical techniques struggle to fully capture the complex context within the data, limiting their ability to produce diverse patterns that can enhance model performance.

Recently, there has been a shift towards generating CAD using pre-trained language models (PLMs)~\citep{Madaan_Padhi_Panwar_Saha_2021, wu-etal-2021-polyjuice,zhou-etal-2022-flipda, wen-etal-2022-autocad, dixit-etal-2022-core, liu-etal-2022-wanli, chen-etal-2023-disco}. 
These models offer the advantage of automatically producing high-quality data by reflecting the context of the input data, thereby reducing the reliance on manual methods. However, since PLMs are trained based on the distribution of the training data, they can easily become biased toward frequently occurring contextual patterns.
This causes PLMs to rely more on common patterns, rather than capturing complex sentence structures or uncovering hidden causal relationships.
Moreover, most current approaches focus solely on data augmentation, neglecting the need to enhance model robustness during training.


To improve model robustness while reducing shortcuts, it is necessary not only to focus on contextual patterns but also to learn structural patterns where shortcuts occur.
Models that learn these structural patterns can capture the fundamental meaning of sentences and the relationships between key components, rather than relying on word frequency or specific terms. By doing so, the model effectively avoids shortcuts and achieves stronger generalization. Structural patterns, particularly those involving the roles of subjects, verbs, and objects, remain consistent across different contexts, allowing models to perform more reliably across diverse datasets.

Therefore, the model should learn the causal relationships within sentences to effectively learn structural patterns, rather than just word frequency or simple textual features. This allows the model to focus on important words (nouns, verbs, adjectives, etc.) to grasp the core meaning of the sentence while reducing its reliance on less significant words (pronouns, conjunctions, etc.). Consequently, models that learn structural patterns can provide consistent performance across a variety of sentence structures.

In this paper, we propose an effective method, \textbf{SALAD}, for enhancing model robustness while addressing spurious correlations, called 
\textbf{S}tructure \textbf{A}ware and \textbf{L}LM-driven \textbf{A}ugmented \textbf{D}ata for Contrastive Learning.
This method combines \textit{structure-aware augmented data} (positive) generated by a tagging-based approach that considers structural patterns and \textit{counterfactually augmented data} (negative) generated by large language models (LLMs) to create complex and diverse sentence patterns.
Then, contrastive learning is applied to enable the model not only to classify patterns but also to capture relationships between samples.
Using the original sentence as the anchor, the positive sample is trained to remain close to it while maintaining the structural patterns where shortcuts occur, whereas the negative sample, generated by the LLM, is trained to diverge from the anchor by incorporating diverse patterns.
Our \textit{SALAD} helps the model effectively learn key patterns in simple sentence structures, reducing its dependence on shortcuts.
Simultaneously, the model learns negative patterns in various contexts, enhancing its generalization performance. 
We conducted experiments on three tasks: Sentiment Classification, Sexism, and Natural Language Inference, to verify the model's robustness across various environments. Furthermore, we demonstrated our model's generalization performance in out-of-distribution datasets and cross-domain settings.

\begin{figure*}[t]
\centering
  \includegraphics[width=\textwidth]{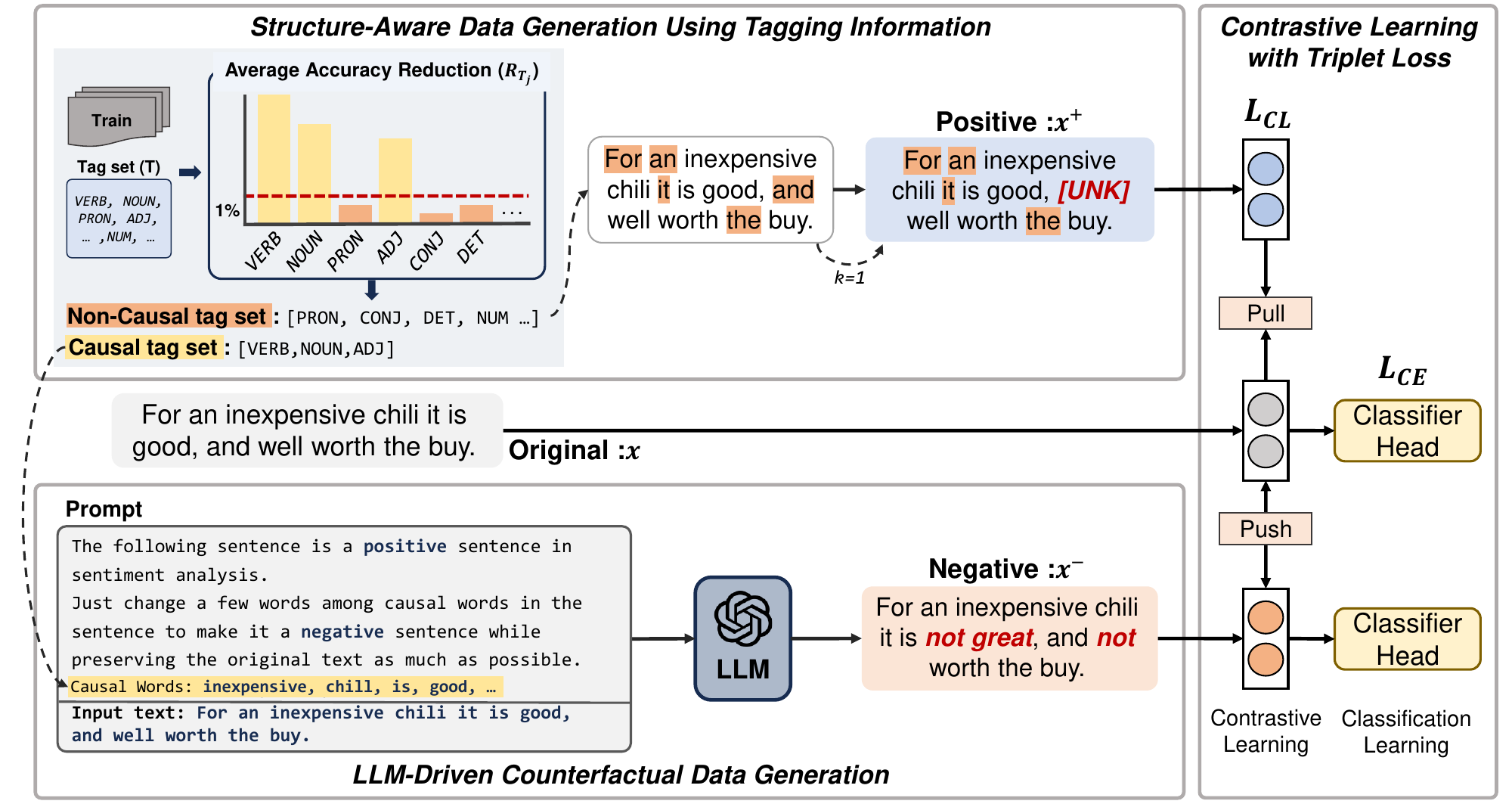}
  \vspace{-0.7cm}
  \caption{\textbf{Overview of \textit{SALAD}.} Our proposed method consists of three steps. First, we use a tagging-based method to generate positive data based on the structure where shortcuts occur (Sec.~\ref{3.1}). Next, we use an LLM to generate counterfactual data to capture complex and diverse sentence patterns (Sec.~\ref{3.2}). Finally, contrastive learning is applied to effectively capture key sentence structural patterns between our augmented data, minimizing spurious correlations and enhancing generalization performance (Sec.~\ref{3.3}).}
  \label{figure:fig1}
\end{figure*}

\section{Related work}

\paragraph{Text Data Augmentation}
Data augmentation has traditionally been used to improve model performance and increase data diversity. 
EDA~\cite{wei-zou-2019-eda} applied simple heuristic transformations such as synonym replacement, word insertion, deletion, and swapping. While it was easy to implement and low-cost, it has the limitation of not considering context.
PLM-based methods, such as SSMBA~\cite{ng-etal-2020-ssmba},
used pre-trained models like BERT~\cite{devlin-etal-2019-bert} to corrupt and reconstruct data, improving performance, particularly on out-of-domain datasets. 
However, this approach requires significant computational resources and could introduce noise. 
LLM-based methods, like AugGPT~\cite{dai2023auggpt}, leveraged GPT-3 to rephrase sentences and enhance performance in few-shot learning settings, but they relied heavily on large language models, which were resource-intensive. While these methods contributed to increased data diversity, they had the drawback of lacking fine-grained control over the quality and relevance of the generated data.

\paragraph{Counterfactual Data Augmentation}
Counterfactual augmented data has been generated using various approaches.
Early methods relied heavily on manual annotation, where human annotators made minimal changes to flip the label of the original text~\cite{Kaushik2020Learning, samory2021call}.
This manual approach was effective in improving the robustness and generalization of text classification models but was also time-consuming and costly.
To address the limitations of manual annotation, rule-based methods were introduced. These methods employed fixed rules such as sentiment dictionaries~\cite{yang-etal-2021-exploring} or named-entity tags, semantic role labels, and sentiment information~\cite{Madaan_Padhi_Panwar_Saha_2021} to automatically generate CAD.
However, these approaches were restricted by the rigid nature of predefined rules, limiting the quality and flexibility of the generated data.
More recently, LLMs have been explored for CAD generation. For example, GPT-3 has been combined with counterfactual retrievers to automatically produce CAD~\cite{dixit-etal-2022-core}, while other frameworks have utilized collaboration between human workers and LLMs to create effective datasets~\cite{liu-etal-2022-wanli}.
Additionally, GPT-3 has been employed to generate high-quality data for natural language inference (NLI) tasks~\cite{chen-etal-2023-disco}. Despite these advancements, many LLM-based methods still require human validation or the use of additional models for data verification. Furthermore, the majority of these studies have focused on data augmentation without addressing practical training strategies aimed at improving model robustness.

\paragraph{Robust Learning}
Recent methods for improving robustness in text classification have explored a range of strategies. One approach combines cross-entropy loss with SupCon loss~\cite{khosla2020supervised} which has shown improvements in general performance and robustness but does not fully address spurious correlations~\cite{gunel2021supervised}.
To tackle the spurious correlation problem, some methods focus on identifying and removing shortcut-related features. For example, matched sample techniques have been used to distinguish between shortcuts and genuine patterns, improving robustness by filtering out shortcut-related words~\cite{wang-culotta-2020-identifying}. Further methods include using cross-domain analysis and knowledge-aware perturbations to differentiate spurious tokens from important ones~\cite{wang-etal-2022-identifying}.
Another approach is causally contrastive learning, which trains models to identify causal features, improving their robustness against spurious correlations~\cite{Choi_Jeong_Han_Hwang_2022}. Despite these advancements, many of these methods still rely on gradient-based techniques and fine-tuned classifiers, which can be biased themselves, limiting their effectiveness in fully overcoming spurious correlations.

\section{Proposed Method}

Figure \ref{figure:fig1} illustrates an overview of \textit{SALAD}, consisting of the following three processes: Structure-aware data generation, LLM-driven counterfactual data generation, and contrastive learning with triplet loss. 

\subsection{Structure-Aware Data Generation Using Tagging Information} \label{3.1}
We propose a method to construct positive data that reduces bias from non-causal words and spurious correlations, enhancing model robustness and generalization. 
To address these issues, it is crucial that the model learns structural patterns where shortcuts occur by focusing on understanding the causal relationships within sentences, rather than relying on word frequency or simple textual features.
Therefore, we leverage the Part-of-Speech (POS)~\citep{petrov-etal-2012-universal} tag set to identify structural information where shortcuts are likely to occur. Then, we construct a non-causal POS tag set $G$.

Given a collection of training data $D = {\{x_i\}}^m_{i=1}$ and the universal POS tag set $T = \{\textit{VERB}, \textit{NOUN}, \dots, \textit{DET}\}$, we remove all tokens corresponding to each POS tag $T_j$. For example, by removing all words corresponding to \textit{VERB} from the data, we construct $D_{\backslash \textit{VERB}} = {\{o_i\}}^m_{i=1}$. We then calculate the average accuracy reduction from the standard fine-tuned model $f$ to determine which tags are irrelevant to the label.
\begin{equation}
   R_{T_{j}}=\frac{1}{m}\sum_{i=1}^{m}(f(x_i)-f(o_{i}))
   \label{equation:importance}
\end{equation}

If the average accuracy reduction $R_{T_{j}}$ exceeds a \textit{threshold}, we consider that $T_j$ is a causal POS tag, meaning it is directly associated with the label. Conversely, if the reduction is smaller than the \textit{threshold}, we assume that $T_j$ belongs to the non-causal POS tag set, indicating it does not affect the label.


After defining the non-causal POS tag set $G$, we randomly select $k$ tokens from each sentence $\{x_i\}$ that belong to $G$ and replace these tokens with the \verb|[UNK]| token, constructing a set of structure-aware positive data, denoted as $D_{pos}={\{x^{+}_i\}}^m_{i=1}$.
This allows us to focus on the genuine tokens that influence the label while ignoring the tokens involved in the structural pattern where the shortcut occurs.
Here, $k$ is determined by multiplying a scaling factor $\alpha$ that reflects the average number of non-casual words in the training data. 

As shown in Figure~\ref{figure:fig1}, words such as `For', `an', and `the' can be identified as non-causal words.
The key point is that our method considers different non-causal words at each epoch while maintaining structural information that does not affect the label. This helps mitigate the tendency for the model to become biased towards spurious correlations as training progresses.

\subsection{LLM-Driven Counterfactual Data Generation}~\label{3.2}
To enhance model robustness and improve generalization, it is essential to incorporate not only structure-aware data but also diverse sentence patterns.
A key aspect of our proposed method is leveraging LLMs to generate counterfactually negative data.
Although the LLM-generated counterfactual sentences involve only minor changes prompted by simple instructions, they provide sufficient contextual variation and offer a range of sentence patterns.


Given a collection of training data ${\{x_i\}}^m_{i=1}$, we construct counterfactual data, denoted as $D_{neg}={\{x^{-}_i}\}^m_{i=1}$, using LLM. In contrast to recent studies using LLMs to generate counterfactually augmented data~\citep{dixit-etal-2022-core, liu-etal-2022-wanli, chen-etal-2023-disco}, we focus on generating counterfactual data using a simple prompt. 
To achieve this, we provide additional word information corresponding to the causal POS tag set in the prompt, minimally altering genuine tokens that directly impact the label.
The causal POS tag set is obtained through the same process as deriving the non-causal tag set in Section~\ref{3.1}, defining the causal POS tag set if their accuracy reduction exceeds the \textit{threshold}.

Appendix~\ref{appendixb} provides instructions used in our method. For the sentiment and sexism task, negative samples are created by changing tokens such that the label flips from \textit{Positive (or sexist)} to \textit{Negative (or non-sexist)} or from \textit{Negative (or non-sexist)} to \textit{Positive (or sexist)}. For the NLI task, we only consider two cases for generating negative samples: \textit{Entailment} to \textit{Contradiction} and \textit{Contradiction} to \textit{Entailment}, excluding the \textit{Neutral}.


\subsection{Contrastive Learning with Triplet Loss}~\label{3.3}
Finally, we use contrastive learning for effective training of models in the generated counterfactual and positive data.
First, the counterfactual data generated by altering only genuine tokens using LLM are considered not only to be a loss for direct label prediction but also to be a loss that encourages them to move further away from the original sentence in the latent space.
Next, by bringing the positive data (structure-aware augmented data) closer to the original samples in the representation space, we effectively mitigate bias towards non-causal words and enhance the model's generalization ability.
In summary, we aim to emphasize important features and eliminate unnecessary shortcuts through the generated triplets.
In conventional fine-tuning models, the \verb|[CLS]| hidden representations from PLM are passed through a classifier head to produce the probability distribution over the label set $y$. As a result, the model parameters $\theta$ are trained to minimize the cross-entropy loss between the predicted label $\hat{y}$ and the ground-truth label $y$:
\begin{equation}
    L_{CE} = \sum_{i=1}^{N}\sum_{c=1}^{C}y_{i,c}\cdot \log\hat{y}_{i,c}
    \label{equation:CE}
\end{equation}
where $N$ denotes a batch of training examples of size and $C$ denotes classes.


We utilize a loss function similar to the training approach in C2L~\citep{Choi_Jeong_Han_Hwang_2022}, which applied a margin-based ranking loss. The specific calculation of the triplet loss is as follows:

\begin{equation}
\begin{aligned}
    L_{CL} = \max(0, \qquad \qquad \qquad \qquad \qquad\\
    {1 \over m} \sum_{i=1}^{m} d(x_{i}, x^{+}_{i}) - d(x_{i}, x^{-}_{i}) + \alpha)
\end{aligned}
\end{equation}
where $m$ is the number of sentences, $x_{i}$ represents the $i$-th original sentence, $x^{-}_{i}$ and $x^{+}_{i}$ are the negative and positive sentence generated by our method, respectively. $\alpha$ is a margin value enforced between positive and negative pairs, and $d(\cdot)$ computes the distance between the hidden states of \verb|[CLS]| tokens as the representations of two sentences.

The final loss function, which combines both the classification objective and the contrastive learning objective, is as follows:
\begin{equation}
    L = (1-\lambda)L_{CE} + \lambda L_{CL}
    \label{equation:total_loss}
\end{equation}
where $\lambda$ is a scalar weighting hyperparameter that balances the two loss components and is tuned separately for each downstream task.

\section{Experiment Setups}

\subsection{Datasets}
\paragraph{Source Datasets}
To validate the ability of our method to address the phenomenon of being biased by spurious correlation in training data in various tasks, we conduct experiments on three tasks: sentiment classification, sexism classification, and natural language inference.

For sentiment classification and natural language inference tasks, we use the original dataset from \citet{Kaushik2020Learning} where a counterfactually-revised dataset (CF) is paired with the original dataset (O).
For the sexism classification task, we use one origin dataset from \citet{samory2021call}, where the dataset contains pairs annotated by crowd workers, where sexist sentences are revised to non-sexist counterparts. 
We use these original-counterfactual pairs and ensure label balance by constructing an additional non-sexist dataset sampled from non-pairs in the dataset. 
Further, the original dataset is split in a 9:1 ratio for training and testing, with 10\% of the training dataset aside for validation.

\paragraph{Evaluation Datasets}
In all tasks, we evaluate the robustness and generalization ability of our method using three in-domain datasets (IDD) and out-of-distribution datasets (ODD). 
For IDD evaluation, we use the test set from the original dataset (O-test) and the test set from the counterfactual dataset (CF-test). 
In the NLI task, the CF-Test datasets utilize a revised counterfactual dataset that combines both premise-revised and hypothesis-revised data.

For ODD evaluation, we use YELP~\citep{DBLP:journals/corr/Asghar16}, SST2~\citep{socher-etal-2013-recursive}, FineFood~\citep{mcauley2013amateurs}, and the 
Tweet~\footnote{https://www.kaggle.com/c/tweet-sentiment-extraction} for sentiment analysis. 
For the sexism task, we use another Tweet~\footnote{https://www.kaggle.com/datasets/dgrosz/sexist-workplace-statements}. 
For the NLI task, we use the two types of MNLI~\citep{mnli} dataset.


\paragraph{Cross-Domain Datasets}
To further demonstrate the generalization ability of our method, we conduct cross-domain experiments. We use three sentiment datasets on SST-2~\citep{socher-etal-2013-recursive}, IMDb~\citep{maas-etal-2011-learning}, FineFood \cite{mcauley2013amateurs}. We utilize official train, validation, and test sets if available. In cases where such datasets are not provided, we randomly split the data into training and validation sets with an 8:2 ratio for each seed.

\begin{table*}[t]
\centering
\resizebox{0.9\textwidth}{!}{
\begin{tabular}{l|cc|cccc|c}
\toprule
\multirow{2}{*}{\textbf{Methods}} &  \multicolumn{2}{c|}{\textbf{In-Domain Dataset}} & \multicolumn{4}{c|}{\textbf{Out-of-Distribution Dataset}} & \multirow{2}{*}{\textbf{Overall}} \\  \cmidrule(lr){2-7} 
& O-Test & CF-Test & YELP & SST2 & FindFood & Tweet  \\
\midrule
\textbf{\textit{Standard Fine-Tuning (full-data)}} & & & & & & &\\
RoBERTa-large~\citep{liu2019roberta} & \textbf{94.13} & 92.28 & 94.85 & 79.41 & 95.24 & 73.04 & 88.16 \\
\midrule
\textbf{\textit{Robust Learning}} & & & & & & &\\
SupCon \cite{gunel2021supervised} & {\ul 93.85}	&88.11	&95.26	&86.20	&95.32	&74.90	& 88.94 \\ 
C2L~\citep{Choi_Jeong_Han_Hwang_2022} & 93.37 & 93.03 &93.19	&79.90	&94.26	&68.85	&87.10 \\ \midrule
\textbf{\textit{Text Data Augmentation}} & & & & & & & \\ 
EDA~\citep{wei-zou-2019-eda} & 93.58 & 93.72 & 95.28 & 89.73 & 95.40 & 81.24 & {\ul 91.49} \\
SSMBA~\citep{ng-etal-2020-ssmba} & 93.60 & 92.69 & \textbf{95.90} & 89.40	& \textbf{96.12} & 78.75	& 91.08 \\
AugGPT~\citep{dai2023auggpt} & 93.37 & 91.46	& {\ul 95.32} & {\ul 90.21} & 94.18 & 78.66 & 90.53 \\ \midrule
\textbf{\textit{Counterfactual Data Augmentation}} & & & & & & &\\
Human-CAD~\citep{Kaushik2020Learning} & 93.17	&{\ul 95.47}	&92.16	&88.65	&94.26	&80.66	&90.73 \\ 
CORE-CAD~\citep{dixit-etal-2022-core} & 91.73	&95.15	&89.70	&90.10	&93.06	&\textbf{86.77}	&91.09 \\
\midrule
\textbf{SALAD} & 93.78	&\textbf{95.90}	&94.99	&\textbf{92.68}	&{\ul 95.58}	&{\ul 85.35}	&\textbf{93.05} \\
\bottomrule
\end{tabular}}
\caption{\textbf{Accuracy of various approaches in sentiment classification task:} For the in-domain dataset, we use the original test set (O-Test) and counterfactual test set (CF-Test). The best performance is highlighted in \textbf{boldface}, and the second-best is marked as \underline{underlined}.}
\label{table:Main_results_sentiment}
\end{table*}

\begin{table}[t]
\centering
\resizebox{\linewidth}{!}{
\begin{tabular}{l|cc|c|c}
\toprule
\multirow{2}{*}{\textbf{Methods}} & \multicolumn{2}{c|}{\textbf{IDD}} & \textbf{ODD} & \multirow{2}{*}{\textbf{Overall}} \\ \cmidrule{2-4}
 & O-Test & CF-Test  & Tweet &  \\ \midrule
RoBERTa-large & 92.69 & 49.23  & 81.00 & 72.49 \\ \midrule
SupCon & 91.79 & 22.56  & 76.28 & 60.84 \\
C2L & \textbf{93.21} & 37.69  & 77.92 & 67.18 \\ \midrule
EDA & 91.67 & 37.69  & 81.59 & 67.74\\
SSMBA & 92.82 & 25.64  & 79.36 & 63.02 \\
AugGPT & 92.31 & 29.23 & 78.83 & 64.08 \\ \midrule
Human-CAD & 91.79 & \textbf{91.80} & {\ul 83.11} & \textbf{89.47} \\ \midrule
\textbf{SALAD} & {\ul 93.07}	&{\ul 88.47}	&\textbf{83.38}	&{\ul 88.31} \\ \bottomrule
\end{tabular}}
\caption{\textbf{Accuracy of various approaches in sexism classification task:} We evaluate on in-domain (IDD) and out-of-distribution (ODD) dataset.}
\label{table:Main_results_sexism}
\end{table}

\subsection{Baselines}
We compare our proposed method with various data augmentation techniques, text data augmentation methods~\citep{wei-zou-2019-eda, ng-etal-2020-ssmba, dai2023auggpt}, counterfactual data augmentation methods~\citep{Kaushik2020Learning, dixit-etal-2022-core, chen-etal-2023-disco}, and robust learning methods~\citep{gunel2021supervised, Choi_Jeong_Han_Hwang_2022}. Detailed explanations are provided in the Appendix \ref{appendix:baselines}.

\subsection{Implementation Details}
For all experiments, we use RoBERTa-large~\citep{liu2019roberta} as our PLM backbone, the batch size is 16, and the learning rate is 1e-05. For sentiment and sexism classification tasks, the maximum sequence length is 256, while it is 128 for NLI tasks. Also, we run all experiments three times with three different random seeds and report the average performances. 
For each experiment that includes a contrastive objective, we employ different scalar weighting hyperparameters $\lambda$ for each dataset that achieves the best performance. For counterfactual sample generation, we use GPT-4o-mini from OpenAI with a temperature of 0.1 and a Top-p value of 1. The \textit{threshold}, used for determining the causal and non-causal tag set $G$, 0.1 across all datasets. The parameter $\alpha$ is defined as \texttt{0.18}, and based on the value of $\alpha$, $k$ is defined differently for each dataset. 
The dataset statistics, hyperparameters for each dataset, and the non-causal tag sets used in constructing positive samples are presented in Appendix~\ref{appendixa}.



\section{Results}

\begin{table*}[t]
\centering
\resizebox{0.8\textwidth}{!}{
\begin{tabular}{l|cc|cc|c}
\toprule
\multirow{2}{*}{\textbf{Methods}} & \multicolumn{2}{c}{\textbf{In-Domain}} & \multicolumn{2}{|c|}{\textbf{Out-of-Distribution}} & \multirow{2}{*}{\textbf{Overall}} \\ \cmidrule(lr){2-5}
 & O-test & CF-test &  MNLI\textsuperscript{1} & MNLI\textsuperscript{2}  \\ \midrule
\textit{\textbf{Standard Fine-Tuning (full-data)}}  &  &  &  &  &  \\
RoBERTa-large \cite{liu2019roberta} & 87.50	&69.90 &73.27	&73.97	&{\ul 76.16} \\ \midrule
\textit{\textbf{Robust Learning}} &  &  &  &  &  \\
SupCon \cite{gunel2021supervised} & 86.42	&60.03	&64.70	&64.39	&68.89 \\
C2L \cite{Choi_Jeong_Han_Hwang_2022} & 87.96	&68.49	&72.18	&72.74	&75.34 \\ \midrule
\textit{\textbf{Text Data Augmentation}} &  &  &  &  &  \\
EDA \cite{wei-zou-2019-eda} & 86.59	&67.58	&70.93	&71.12	&74.06 \\
SSMBA \cite{ng-etal-2020-ssmba} & 87.16	&63.54		&72.03	&72.95	&73.92 \\
AugGPT \cite{dai2023auggpt} & 86.92	&69.61		&{\ul 73.62}	&{\ul 74.38}	&76.13 \\ \midrule
\textit{\textbf{Counterfactual Data Augmentation}} &  &  &  &  &  \\
Human-CAD~\citep{Kaushik2020Learning} & {\ul 88.25}	&71.60		&71.74	&71.47	&75.76 \\
CORE-CAD~\citep{dixit-etal-2022-core} & 64.65	&57.26		&62.60	&62.98	&61.88 \\ 
DISCO \cite{chen-etal-2023-disco} & 79.84	&{\ul 78.66}		&68.42	&67.60	&73.63 \\ \midrule
\textbf{SALAD} & \textbf{88.40}	&\textbf{80.91}	&\textbf{74.06}	&\textbf{74.93}	&\textbf{79.57} \\ \bottomrule
\end{tabular}}
\caption{\textbf{Accuracy of various approaches in natural language inference task:} For out-of-distribution dataset, MNLI\textsuperscript{1} refers to MNLI-hard-match, and MNLI\textsuperscript{2} refers to MNLI-hard-mismatch. The best performance is highlighted in \textbf{boldface}, and the second-best is marked as \underline{underlined}.}
\label{table:Main_results_nli}
\end{table*}

\begin{table*}[t]
\centering
\resizebox{0.9\textwidth}{!}{
\begin{tabular}{l|cc|cc|cc|c}
\toprule
\textbf{Methods} & S $\rightarrow$ I & S $\rightarrow$ F & I $\rightarrow$ S & I $\rightarrow$ F & F $\rightarrow$ S & F $\rightarrow$ I & \textbf{Overall}  \\
\midrule
\textbf{\textit{Standard Fine-Tuning (full-data)}} & & & & & & \\
RoBERTa-large \cite{liu2019roberta} & {\ul 91.67} & 93.08  & 89.16 & 91.13  & \underline{82.48} & 90.22 & 89.62  \\
\midrule
\textbf{\textit{Robust Learning}} & & & & & & &\\
SupCon \cite{gunel2021supervised} & 90.82 & 89.64 & \underline{91.21} & \underline{94.95} & 73.40 & 89.68 & 88.28 \\ 
C2L \cite{Choi_Jeong_Han_Hwang_2022} & 90.52 & 91.61 & 89.90 & 94.64 & 81.18 & {\ul 90.50} & 89.72\\
\midrule
\textbf{\textit{Text Data Augmentation}} & & & & & & &\\
EDA \cite{wei-zou-2019-eda} & 91.64 & \underline{93.51} & 90.76 & 94.12 & 80.18 & 89.29 & \underline{89.92} \\ 
SSMBA \cite{ng-etal-2020-ssmba} & 90.71 & 90.78 & \textbf{94.21} &  93.96 & 78.75 & 89.31 & 89.62\\
\midrule
\textbf{SALAD} & \textbf{92.41}	& \textbf{94.19}  & 90.88 & \textbf{94.96} & \textbf{86.00}	& \textbf{91.25} & \textbf{91.61}\\
\bottomrule
\end{tabular}}
\caption{\textbf{Accuracy of cross-domain task:} We evaluate across three datasets, SST-2 (S), IMDB (I), and FineFood (F) to evaluate the model's generalization ability of our method.}
\label{table:cross_domain_result}
\end{table*}

In this section, we demonstrate the outperformance of \textit{SALAD} for the robustness and generalization abilities of the model in three tasks. We also conduct comprehensive ablation studies to demonstrate its superiority.


\subsection{Main Result}

\begin{table*}[t]
\centering
\small
\resizebox{0.8\textwidth}{!}{
\begin{tabular}{l|cccc|cc|cc}
\toprule
\multirow{2}{*}{Methods} & \multicolumn{2}{c}{Data Augmentation} & \multicolumn{2}{c}{Loss} & \multicolumn{2}{|c}{Sentiment Task} & \multicolumn{2}{|c}{NLI Task} \\ \cmidrule{2-9} 
 & Neg & Pos & CE & CL & IDD & ODD & IDD & ODD \\ \midrule
Human-CAD & Human & X & X & X & 94.32 & 88.93 & {\ul 79.93} & 71.60 \\
CORE-CAD & GPT & X & X & X & 93.44	& 89.91 & 60.96	& 62.79\\
GPT-CAD & GPT (ours) & X & X & X & 92.42 & 89.59 & 71.35 & 68.78 \\ \midrule
\textbf{SALAD-EDA} & GPT (ours) & EDA & O & O & \textbf{94.98}	&91.45	&79.05	&{\ul 73.84} \\
\textbf{SALAD-GPT} & GPT (ours) & AugGPT & O & O & {\ul 94.88}	&{\ul 91.86}	&78.70	&72.51 \\
\textbf{SALAD} & GPT (ours) & PosTag & O & O & 94.84 & \textbf{92.15} & \textbf{84.65}	& \textbf{74.49} \\ \bottomrule
\end{tabular}}
\caption{\textbf{Accuracy based on variations in SALAD:} GPT (ours) refers to counterfactually augmented data generated by GPT-3.5 using a simple prompt, and PosTag is generated structure-aware augmented data using pos-tagging information. IDD represents the in-domain dataset, and ODD represents the out-of-distribution dataset.}
\label{table:ablation_study}
\end{table*}


\paragraph{Robustness and Generalization}

As shown in Table \ref{table:Main_results_sentiment}, \ref{table:Main_results_sexism}, and \ref{table:Main_results_nli}, \textit{SALAD} demonstrates improved overall accuracy compared to \textit{standard fine-tuning baseline} across all tasks. While it shows a slight decrease of about 1.16\% compared to \textit{Human-CAD} in the sexism classification task, making it the second-best, \textit{SALAD} achieves the best performance in all other tasks. This indicates that our method is robust and performs similarly to human-generated high-quality CAD, even surpassing it in the sentiment classification task.

In particular, the results of the CF-Test for the sexism classification task show that \textit{SALAD}'s performance drop is relatively small at just 3.33\%, while other baselines show a significant performance drop compared to the CF-Test, indicating a lack of robustness. This suggests that fine-tuned PLMs using our \textit{SALAD} are less sensitive to spurious patterns.

Furthermore, \textit{SALAD} is particularly effective in ODD scenarios. 
While \textit{Human-CAD} achieves the highest performance in IDD due to the use of CF-Train during training, its performance on ODD is consistently lower compared to \textit{SALAD}. This underscores that the proposed method significantly enhances generalization capabilities and ensures model robustness, leading to a dramatic improvement in overall performance.


\paragraph{Cross-Domain Generalization}
We additionally experiment with the performance of the domain generalization task to demonstrate that our proposed method is effective in securing robustness and enhancing generalization capabilities. As shown in Table~\ref{table:cross_domain_result}, there is a substantial increase in performance, with the overall accuracy reaching 91.61\% in six cross-domain settings. 

Specifically, except in IMDB (I) $\rightarrow$ SST2 (S), all accuracy achieves the best performance. This indicates that the efforts to address spurious correlations in \textit{SALAD} can potentially contribute to improving generalization abilities, even when the domain undergoes a shift.


\subsection{Effectiveness of Our Data Augmentation}
We conduct ablation studies on \textit{SALAD} for sentiment classification and NLI tasks, focusing on two key aspects: whether our simple negative sample generation method using LLM outperforms using other complex methods and whether our tagging-based positive data augmentation method helps the model effectively learn structural patterns during contrastive learning, thus addressing spurious correlation issues and improving generalization in ODD.



In Table~\ref{table:ablation_study}, the first three rows present the experimental results for training the model using only negative sample augmentation. Here, \textit{GPT-CAD} shows effective performance in both tasks, particularly similar performance with the \textit{Human-CAD}.

The last three rows demonstrate the results of comparing our tagging-based positive data augmentation method with other methods during contrastive learning. \textit{SALAD} outperforms other baselines in both tasks, achieving notably higher performance in ODD. This indicates that our approach effectively enables the model to learn structural patterns, reduce bias, and make more accurate predictions in ODD scenarios, free from spurious correlations.

\begin{table}[t]
\centering
\setlength{\tabcolsep}{0.5em} 
\small
\begin{tabular}{l|ccc}
\toprule
 & \textbf{Diversity} & \textbf{Overlap (\%)} & \textbf{BERTScore}\\
\midrule
Human-CAD & 1,392 & 92.68  &0.969\\
CORE-CAD& 498 & 60.15  &0.914\\
\midrule
\textbf{GPT-CAD} & 1,490 & 86.77 & 0.969\\
\bottomrule
\end{tabular}
\caption{Analysis of the method of CAD on sentiment analysis. GPT-CAD is a counterfactually augmented dataset proposed in \textit{SALAD}.}
\label{table:qual_cd_1}
\end{table}

\begin{table}[t]
\small
    \centering
        \begin{tabular}{p{7.3cm}}
            \toprule
            \textbf{Original Sentence}  \\
            Long, boring, blasphemous. Never have I been so glad to see ending credits roll. \\
            \midrule
            \midrule
            \textbf{Human-CAD} \\
            Long, \textcolor{violet}{fascinating}, \textcolor{violet}{soulful}. Never have I been so \textcolor{violet}{sad} to see ending credits roll.\\
            \midrule
            \textbf{CORE-CAD} \\
            \textcolor{violet}{I don't know why I hate this movie so much, now I am tired of watching it.}\\
            \midrule
            \textbf{GPT-CAD} \\
            \textcolor{violet}{Short}, \textcolor{violet}{exciting}, \textcolor{violet}{delightful}. \textcolor{violet}{Always} have I been so \textcolor{violet}{happy} to see \textcolor{violet}{the beginnin}g credits roll. \\
            \bottomrule
        \end{tabular}
    \caption{Example of counterfactual data of our method (\textit{GPT-CAD}) and other baselines. The \textcolor{violet}{purple} indicates where the tokens are changed to flip the label.}
    \label{table:qual_cd_2}
\end{table}

\subsection{Quality of LLM-Driven Augmented Data}
We evaluate our generated LLM-based counterfactual augmented data (GPT-CAD) in three metrics, as shown in Table \ref{table:qual_cd_1}. 
First, we measure the number of new corpora that did not appear in the original train dataset (\textbf{Diversity}). Second, we calculate the ratio of corpora that overlap with the original train dataset's (\textbf{Overlap}). 
Lastly, to examine how well the generated counterfactual sentences maintain the existing context, we use BERTScore~\cite{Zhang*2020BERTScore:}, which computes cosine similarity between the original sentences and the generated counterfactual sentences using BERT encodings (\textbf{BERTScore}). 
As a result of our evaluation, \textit{GPT-CAD} demonstrates performance close to \textit{Human-CAD}, particularly with a BERTScore and diversity. 
This highlights the effectiveness of our LLM-driven counterfactual data generation compared to manually generated data.
Additionally, as shown in Table~\ref{table:qual_cd_2}, \textit{GPT-CAD} effectively preserves the original sentence structure while changing only a minimal number of genuine tokens, outperforming previous methods in terms of both structure preservation and context. This suggests its ability to preserve the original context while altering keywords.

\section{Conclusion}
In this paper, we introduced \textbf{SALAD},
a method that enhances model robustness and generalization through the use of structure-aware and counterfactually augmented data for contrastive learning. 
By combining a tagging-based method for generating structure-preserving positive samples and LLM-generated counterfactual negative samples, \textit{SALAD} enables models to learn meaningful structural relationships while reducing their reliance on spurious correlations.
Our experiments, conducted on three tasks demonstrate that \textit{SALAD} significantly improves model performance across diverse environments, particularly in out-of-distribution and cross-domain settings. These results highlight the potential of our approach to address the challenges of spurious correlations in natural language processing, providing a more robust and generalizable solution.

\section{Limitation}

In this work, we utilized the GPT-4o-mini model to generate the dataset. GPT-CAD for \textit{SALAD} consists of data in which sentence labels are flipped without requiring human intervention or additional models. While CADs that are re-labeled or generated by humans may yield better performance, our focus is not on meticulously generating CADs. Instead, we aim to verify and analyze the effectiveness of learning with CADs. 
Therefore, in future work, we believe that as higher-quality CADs become available, our proposed framework can be effectively utilized to further enhance model performance.

\section*{Acknowledgments}
This work was partly supported by the Institute of Information \& communications Technology Planning \& Evaluation (IITP) grant and the National Research Foundation of Korea (NRF) grant funded by the Korea government(MSIT) (RS-2019-II190421 (10\%),  No.2022-0-01045 (45\%), No.RS-2024-00360227 (45\%))

\bibliography{anthology, custom}

\clearpage
\appendix

\section{Baselines}
\label{appendix:baselines}
\paragraph{Supcon~\citep{gunel2021supervised}} SupCon is a joint optimization method combining cross-entropy loss and contrastive loss, demonstrating enhanced robustness and improved generalization performance in text classification tasks

\paragraph{C2L~\citep{Choi_Jeong_Han_Hwang_2022}} C2L relies on the classifier model to identify causal words that significantly influence the label to enhance robustness. They treat the masking of causal words as negative examples, and the masking of less significant words as regular positive examples, thereby jointly optimizing triplet loss and cross-entropy. 

\paragraph{EDA~\citep{wei-zou-2019-eda}} This method proposed augmenting sentences by randomly applying four heuristic techniques: synonym replacement, word insertion, word deletion, and word swapping. We employed this method to augment our dataset by applying one augmentation per sentence.

\paragraph{SSMBA~\citep{ng-etal-2020-ssmba}} SSMBA proposed a corrupt-and-reconstruct data augmentation technique using the BERT model, showing performance improvements on out-of-distribution datasets. In our experiments, we adopted the approach of augmenting data while keeping the labels unchanged. We also employed this method to augment our dataset by applying one augmentation per sentence.

\paragraph{AugGPT~\citep{dai2023auggpt}} This method used GPT-3 to augment data, enhancing the performance of text classification in a few-shot setting. In our experiments, we augment data using single-turn dialogues with the prompt ``Please rephrase the following sentence.''

\paragraph{Human-CAD~\citep{Kaushik2020Learning}} This method, which predominantly explores the automated generation of CAD, involves augmenting CAD by human annotators and training it with the original train dataset.

\paragraph{CORE-CAD~\citep{dixit-etal-2022-core}} CORE proposed a retrieval-augmented generation framework for generating CAD using a combination of a retrieval model and GPT-3. In our approach, we use the publicly available dataset on our experimental setup.

\paragraph{DISCO~\citep{chen-etal-2023-disco}} For NLI tasks, we additionally compared DISCO, automatically generating high-quality counterfactual data at scale using the GPT-3 model. 

\section{Implementation Details} \label{appendixa}

\subsection{Experimental Environment}
For all experiments, our experiments are implemented with Pytorch framework~\cite{paszke2019pytorch}, Huggingface trasnformers~\cite{wolf-etal-2020-transformers}, NLTK library \cite{bird-2006-nltk}, OpenPrompt toolkit~\cite{ding2021openprompt}.
We set the environment for all experiments using four NVIDIA 3090 GPUs with 24GB graphic memory, Ubuntu 22.04, Python 3.8, and CUDA 11.7 version.



\subsection{Statistics of Counterfactual Task Dataset}
Table~\ref{table:statistic_CF} shows the statistics of the dataset used in the counterfactual task.

\begin{table}[t]
    \centering
    \begin{subtable}{0.8\linewidth}
        \centering
        \caption*{Sentiment classification task}
        \resizebox{\linewidth}{!}{
        \begin{tabular}{lccc}
            \toprule
            & Positive & Negative & Total \\
            \midrule
            O-Train & 856 & 851 & 1,707 \\
            O-Test & 245 & 243 & 488 \\
            CF-Test & 243 & 245 & 488 \\
            \bottomrule
        \end{tabular}}
    
    \end{subtable}
    \begin{subtable}{\linewidth}
        \centering
        \caption*{Sexism classification task}
        \resizebox{0.8\linewidth}{!}{
        \begin{tabular}{lccc}
            \toprule
            & Sexist & Non-sexist & Total \\
            \midrule
            O-Train & 1,036 & 1,036 & 2,072 \\
            O-Test & 130 & 130 & 260 \\
            CF-Test & 132 & 130 & 262 \\
            \bottomrule
        \end{tabular}}
     \end{subtable}
     \begin{subtable}{\linewidth}
        \centering
        \caption*{Natural language inference task}
        \resizebox{0.9\linewidth}{!}{
        \begin{tabular}{lcccc}
            \toprule
            & Entail & Neutral & Contradict & Total \\
            \midrule
            O-Test & 146 & 123 & 131 & 400 \\
            O-Train & 562 & 554 & 550 & 1,666 \\
            CF-Test & 508 & 554 & 538 & 1,600 \\
            \bottomrule
        \end{tabular}}
        
     \end{subtable}
\caption{Statistics of counterfactual task datasets.}
\label{table:statistic_CF}
     
\end{table}

\subsection{Hyper-parameters}
\begin{figure}[t]
\centering
   \includegraphics[width=\columnwidth]{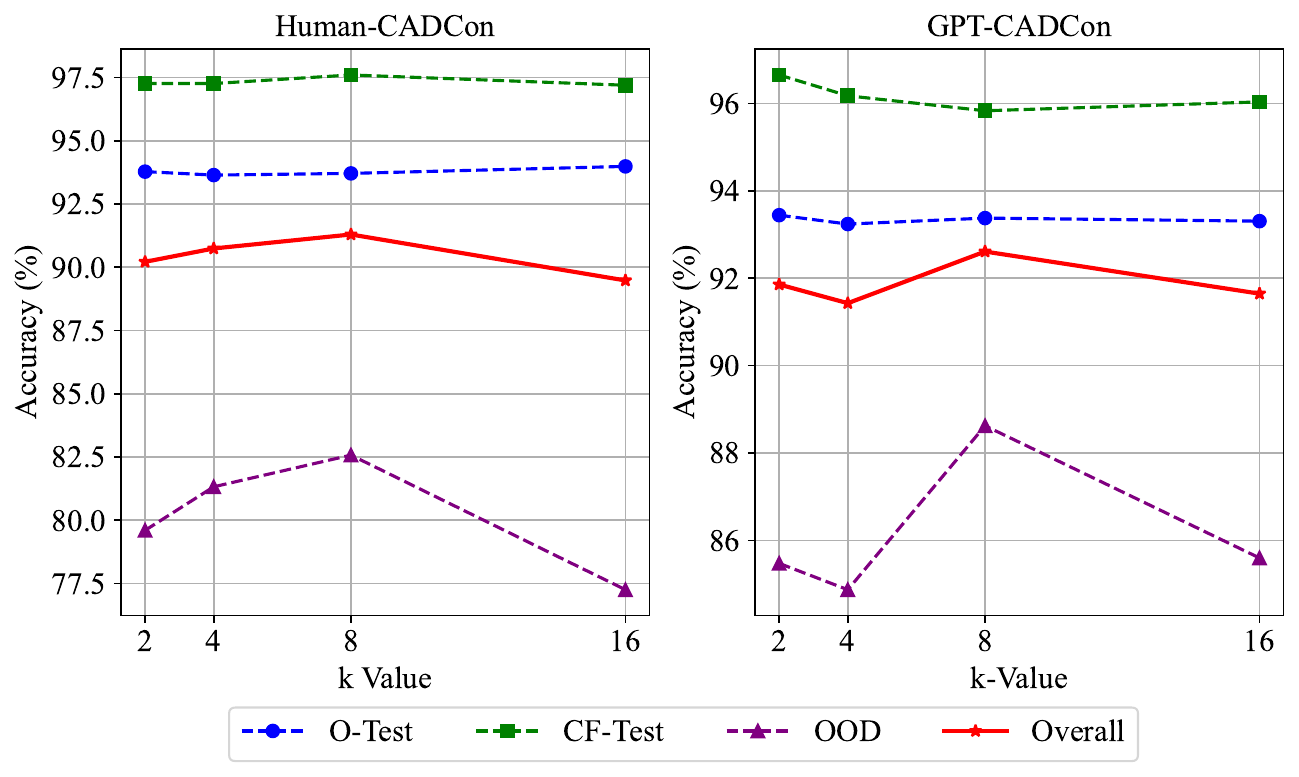}
\caption{\textbf{Experiments on defining $k$:} The value of 8 shows significant performance improvement for the CF-IMDB dataset, particularly on the out-of-distribution dataset (ODD).
}
\label{figure:ablation_k}
\end{figure}

As mentioned in the paper, we employ different hyperparameters, denoted as $k$ and $\lambda$, for each dataset. 
In the structure-aware data generation using tagging information, the parameter $k$ is used to determine the number of word tokens where randomly selected from each sentence that belongs to $G$ and replace these tokens with the \verb|[UNK]| token. According to our experimental results, defining $k$ as 8 showed significant performance improvement for the CF-IMDB dataset, particularly on the out-of-distribution dataset (ODD) (Analysis results are displayed in Figure~\ref{figure:ablation_k}). Therefore, using CF-IMDB as a reference, the scaling factor $\alpha$ was calculated. This calculation is determined by dividing the average number of non-causal tokens, which is 45 for CF-IMDB, resulting in a value of 0.18. Consequently, we calculate the value of $k$ for each dataset by multiplying its respective average non-causal token count with the scaling factor. Summarizing the relevant hyperparameters, they are presented in Table~\ref{table:hyper-parameters}.

\begin{table}[t]
\centering
\small
\begin{tabular}{c|r|r}
\toprule
\multicolumn{1}{c|}{\textbf{Dataset}} & \multicolumn{1}{c|}{\textbf{$k$}} & \multicolumn{1}{c}{\textbf{$\lambda$}} \\
\midrule
CF-IMDB~\citep{Kaushik2020Learning} & 8 & 0.9 \\
CF-NLI~\citep{Kaushik2020Learning} & 1 & 0.9 \\
Sexism \cite{samory2021call}& 1 & 0.3 \\
\midrule
SST2 \cite{socher-etal-2013-recursive}& 1 & 0.1 \\
IMDB \cite{maas-etal-2011-learning}& 8 & 0.9 \\
FineFood \cite{mcauley2013amateurs} & 5 & 0.1 \\
\bottomrule
\end{tabular}
\caption{\textbf{Hyperparameters of \textit{SALAD}:} $k$ represents the number of randomly selected tokens from the non-causal POS tag set ($G$), and $\lambda$ is a scalar weighting hyperparameter used to define the final loss function.}
\label{table:hyper-parameters}
\end{table}

\subsection{Non-Causal Tag Sets Across Datasets}
In the Structure-aware data generation methods, we define the non-causal tag set $G$ by iteratively removing each POS tag set for each dataset and calculating the accuracy reduction. The following Figure \ref{figure:ablation_datasets} is an ablation study on the results of calculating accuracy reductions in sentiment and sexism classification datasets. We estimate $\theta$ to be 1\%, defining the non-causal tag set as the POS information for which the score is less than 1\%. 
The Causal tag sets calculated for each dataset used in our experiment are listed in Table~\ref{table:non-causal-sets}.

\begin{figure}[h!]
\centering
   \includegraphics[width=\columnwidth]{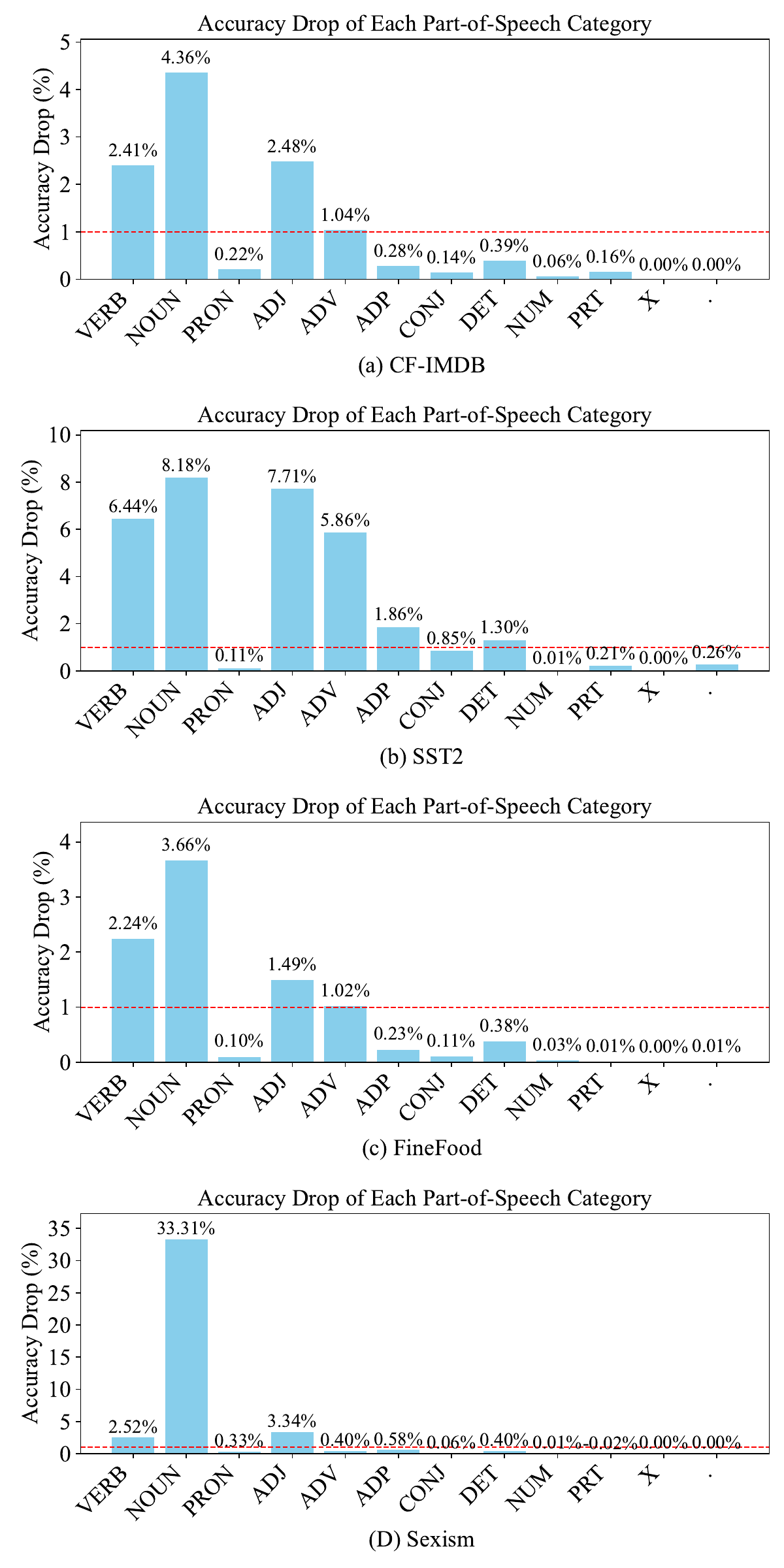}
\caption{
    \textbf{Accuracy reduction of each POS category across datasets:} The $x$-axis represents each POS category, and the $y$-axis represents the average accuracy reduction. We define POS tags with an average accuracy reduction of less than 1\% as the non-causal tag set $G$.
}
\label{figure:ablation_datasets}
\end{figure}

\begin{table}[h]
\centering
\small
\resizebox{\linewidth}{!}{
\begin{tabular}{l|l}
\toprule
\multicolumn{1}{c}{\textbf{Dataset}} & \multicolumn{1}{c}{\textbf{Causal tag set $G$}}  \\
\midrule
CF-IMDB & \textit{VERB, NOUN, ADJ, ADV} \\
CF-NLI & \textit{VERB, NOUN, ADJ, ADV} \\
Sexism & \textit{VERB, NOUN, ADJ}  \\
\midrule
SST2& \textit{VERB, NOUN, ADJ, ADV, ADP, DET} \\
IMDB& \textit{VERB, NOUN, ADJ, ADV} \\
FineFood & \textit{VERB, NOUN, ADJ, ADV}  \\
\bottomrule
\end{tabular}}
\caption{Causal tag sets in each training data.}
\label{table:non-causal-sets}
\end{table}

\section{Analysis of Prompt Instructions} \label{appendixb}
As mentioned in Section~\ref{3.2}, we utilize GPT-4o-mini to generate counterfactual sentences from the original ones using a simple prompt. Since zero-shot LLMs are sensitive to the template used, we conducted experiments with four variations of prompt instructions and selected the most effective one.
\textit{Instruction 1} includes the phrase, ``\textit{Please make it a negative sentence}'', which directly indicates the intended behavior of the model. \textit{Instruction 2} provides the task and label information for the sentence. In \textit{Instruction 3}, we offer more specific guidance with phrases like, ``\textit{Just change a few words}'' and ``\textit{while preserving the original text as much as possible}''. We use a similarly designed prompt, following the format of \textit{Instruction 3}, with adding causal word information, for each task and label. A specific example is shown in Table \ref{table:instructions}.

\begin{table}[t]
\small
  \centering
  \resizebox{\linewidth}{!}{
  \begin{tabularx}{\linewidth}{c X}
    \toprule
    \textbf{Idx} & \textbf{Instructions} \\
    \midrule
    1 & Please make it a negative sentence. \\
    2 & The following sentence is a positive sentence in sentiment analysis. Please make it a negative sentence. \\
    3 & The following sentence is a positive sentence in sentiment analysis. Just change a few words to make it a negative sentence while preserving the original text as much as possible. \\
    4 & The following sentence is a positive sentence in sentiment analysis. Just change a few words among causal words in the sentence to make it a negative sentence while preserving the original text as much as possible. Causal Words: \textbf{\textit{{causal words}}} \\
    \bottomrule
  \end{tabularx}}
  \caption{Examples of instructions for generating negative samples in a sentiment analysis task. We use \textit{Instruction 4} as our prompt to construct counterfactual data in \textit{SALAD}}
  \label{table:instructions}
\end{table}

We aim to compare and analyze the performance and quality associated with each prompt instruction. We evaluate the generated CAD using three metrics. Additionally, we assess the performance of our CAD based on three prompt instructions. \textit{Instruction 1}, which simply flips labels, shows a very low word overlap of 55.26\% with the original sentence. Particularly in \textit{instruction 4}, by incorporating the phrase ``while preserving the original text as much as possible'' and adding causal words information, we identify preservation of up to 86.77\% of the original sentence while flipping the label. Moreover, with a diversity count of 1,490, indicating the number of corpora not used in the original sentence, it can be considered the most superior CAD among the four instructions. The CAD generated with \textit{instruction 4} exhibits similarity to \textit{Human-CAD}, as indicated by the BERTScore.

\begin{table}[t]
\centering
\small
\begin{tabular}{l|ccc}
\toprule
 & \textbf{Diversity} & \textbf{Overlap (\%)} & \textbf{BERTScore} \\
\midrule
Human & 1,392 & 92.68  &0.969 \\
\midrule
Instruction 1 & 758 & 55.26 & 0.895 \\
Instruction 2 & 1,183 & 76.91 & 0.934 \\
Instruction 3 & 1,218 & 83.28 & 0.955 \\
Instruction 4 & 1,490 & 86.77 & 0.969 \\
\bottomrule
\end{tabular}
\caption{Analysis of our generated CAD (GPT-CAD) with different prompt instructions on sentiment analysis.}
\label{table:statistical_CAD}
\end{table} 

\begin{figure}[t]
\centering
   \includegraphics[width=\columnwidth]{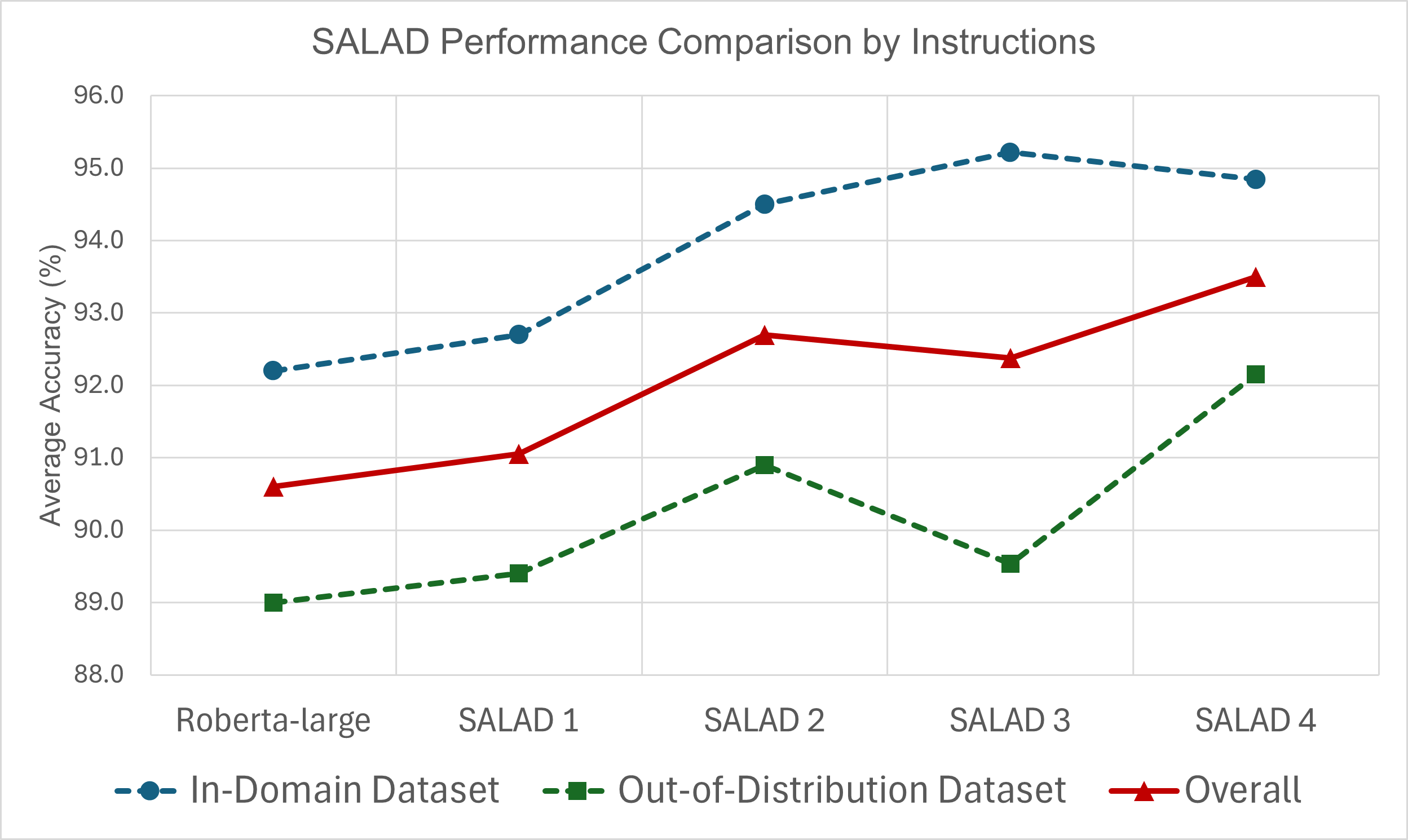}
\caption{
    Performance variations of \textit{SALAD} on datasets generated for each instruction. The number following ``\textit{SALAD}" corresponds to the instructions associated with each number used in Table \ref{table:instructions}.
}
\label{figure:ablation_instructions}
\end{figure}

Also, we conduct an ablation study on datasets generated by four different prompts. Table~\ref{figure:ablation_instructions} reports the performance of applying \textit{SALAD} to the datasets generated through instructions for the four different scenarios. We find that even in instructions where task-related information is limited, such as in \textit{SALAD\_1}, there is a significant improvement in the ability to generalize to ODD data compared to the baseline model Roberta-large. Furthermore, the addition of task-related information in \textit{SALAD\_2} and the inclusion of the instruction ``while preserving the original text as much as possible'' in \textit{SALAD\_3} gradually lead to performance improvements.
Particularly, \textit{SALAD\_4}, which generates CAD with the aim of minimally flipping the label by changing only genuine tokens, proves to be the most effective in achieving robustness through representation learning. 
Consequently, we utilized the \textit{Instruction 4} in all final experiments.

\section{Prompt-based Fine-tuning Settings}

\begin{table*}[t]
\centering
\small
\begin{tabular}{l|cc|cccc|c}
\toprule
\multirow{2}{*}{Methods (8-shot)} &  \multicolumn{2}{c|}{In-Domain Dataset} & \multicolumn{4}{c|}{Out-of-Distribtuion Dataset} & \multirow{2}{*}{Overall} \\ \cmidrule{2-7}  
& O-Test &CF-Test  & YELP & SST2 & Food & Tweet  \\
\midrule
\textbf{\textit{Prompt-based Fine-Tuning}} & & & & & &\\
RoBERTa-large \cite{liu2019roberta} & \underline{92.21} & 90.33 & 93.54  & 82.61 & 94.85 & 72.41 & 88.13 \\
\midrule
\textbf{\textit{Robust Learning}} & &  & & & & &\\
SupCon \cite{gunel2021supervised} & 91.52 & 90.45 & \textbf{95.31} & 84.16 & \underline{95.28} & 73.51 & 88.80 \\ 
\midrule
\textbf{\textit{Data Augmentation}} & & & & & & &\\
EDA \cite{wei-zou-2019-eda} & 91.02 & 91.64 & 94.18 & 84.34 & 94.79 & 71.00 & 88.53 \\ 
SSMBA \cite{ng-etal-2020-ssmba} & \textbf{92.25} & 92.13  & 93.91 & 84.70 & \underline{95.28} & 74.63 & 89.35\\
AugGPT \cite{dai2023auggpt} & 92.13 & 92.30 & 92.68 & 81.55 & 94.64 & 70.53 & 88.07\\
\midrule
\textbf{\textit{Counterfactually Augmented Dataset}} &  & & & & & &\\
Human-CAD & 91.19 & \textbf{93.16} & 94.01 & 85.13 & 94.96 & 78.45 & 90.07 \\ 
CORE-CAD & 91.76 & 92.95  & 93.36 & \underline{88.30} & 93.72 & \underline{81.50} & \underline{90.67}\\
\midrule
\textbf{SALAD} & 91.11 & \underline{91.93}  & \underline{95.28} & \textbf{89.59} & \textbf{95.37}& \textbf{82.23} & \textbf{91.23} \\
\bottomrule
\end{tabular}
\caption{Accuracy of various approaches in sentiment classification task under the prompt-based fine-tuning setting. For the in-domain dataset, we use the original test set (O-Test) and counterfactual test set (CF-Test). The best performance is highlighted in \textbf{boldface}, and the second-best is marked as \underline{underlined}.}
\label{table:fewshot_results}
\end{table*} 

Recently, in order to narrow the gap between pre-training and downstream tasks prompt-based Fine-tuning models are attracting attention and few-shot setting~\cite{NEURIPS2020_1457c0d6, gao-etal-2021-making}. Most prompt-based learning approaches~\cite {shin-etal-2020-autoprompt, schick-schutze-2021-exploiting, gao-etal-2021-making} utilize task-specific templates consisting of discrete prompts alongside input sentences. These prompts contain a \verb|[MASK]| token and are designed to construct an objective that is similar to MLM training, where the goal is to map the \verb|[MASK]| token to the right label (a specific word) with a pre-defined verbalizer. The probability distribution over the label is shown below:

\begin{equation}
    P_M(\verb|[MASK]| = v|T(x))|v \in V_y)
    \label{equation:PT}
\end{equation}

where $T(\cdot)$ is a task-specific template and $V_y$ is the label words of $y$.
We conduct an 8-shot experiment with extremely low data volume for sentiment classification tasks to illustrate the enhancement of robustness.

Table~\ref{table:fewshot_results} presents the results of experiments conducted in prompt-based fine-tuning settings. The results show that \textit{SALAD} achieved state-of-the-art performance with an overall accuracy of 91.23\%. It also delivered the best performance on three out-of-distribution datasets and demonstrated considerable performance in in-domain datasets, achieving the second-best result on the CF-Test.
In conclusion, \textit{SALAD} outperforms in terms of robustness and generalization abilities under the prompt-based fine-tuning setting.

\end{document}